\pdfoutput=1
\documentclass{article}
\usepackage{spconf,amsmath,graphicx,mathtools}
\usepackage{pgfplots}

\title{Single Stream Parallelization of Generalized LSTM-Like RNNs on a GPU}
\name{Kyuyeon Hwang and Wonyong Sung\thanks{This work was supported in part by the Brain Korea 21 Plus Project and the National Research Foundation of Korea (NRF) grants funded by the Ministry of Education, Science and Technology (MEST), Republic of Korea (No. 2012R1A2A2A06047297).}}
\address{Department of Electrical and Computer Engineering\\Seoul National University\\Seoul 151-744, South Korea\\Email: \texttt{khwang@dsp.snu.ac.kr, wysung@snu.ac.kr}}
\begin{document}
\ninept
\maketitle
\begin{abstract}
Recurrent neural networks (RNNs) have shown outstanding performance on processing sequence data. However, they suffer from long training time, which demands parallel implementations of the training procedure. Parallelization of the training algorithms for RNNs are very challenging because internal recurrent paths form dependencies between two different time frames. In this paper, we first propose a generalized graph-based RNN structure that covers the most popular long short-term memory (LSTM) network. Then, we present a parallelization approach that automatically explores parallelisms of arbitrary RNNs by analyzing the graph structure. The experimental results show that the proposed approach shows great speed-up even with a single training stream, and further accelerates the training when combined with multiple parallel training streams.
\end{abstract}
\section{Introduction}
\label{sec:intro}

Deep neural networks have shown quite impressive performances in several pattern recognition applications \cite{hinton2006reducing, hinton2012deep}.  Among the deep neural networks, the feed-forward networks are suitable for processing input data with a fixed length, and they are usually used for image and phoneme recognition.  On the other hand, recurrent neural networks (RNNs) employ feedback inside, and they are suitable for processing input data whose dimension is not fixed or limited. For example, automatic speech recognition (ASR) systems can perform better with an RNN-based language modeling \cite{mikolov2011extensions}.

Since RNNs contain feed-back loops inside, the past input can be memorized and affect the current output. If RNNs are properly trained, it is possible to compress the input history effectively and yield good results even when there are considerable time delays between the input and output.  Especially, the long short-term memory (LSTM) RNN is known to solve the problems with long time lag very successfully \cite{gers2000learning}.  

However, the LSTM RNN employs a very complex component known to be the memory block.  It demands much effort even for slight modification of the structure because of the difficulty in deriving the corresponding training equation. Thus, it is needed to develop a generalized RNN structure that can be modified easily while representing LSTM networks perfectly. Previously, a generalized LSTM-like RNN structure with real-time recurrent learning (RTRL) \cite{williams1989learning} was proposed in \cite{monner2012generalized} with special gated connections. However, we propose a much more general structure by introducing multiplicative layers and delayed connections. Also, we derive a backpropagation through time (BPTT) \cite{werbos1990backpropagation} based training algorithm for our RNN structure, which is generally more flexible than the RTRL-based one.

RNNs also demand very long training time, thus implementation with GPUs or multiprocessors is needed. However, parallelization of the network is difficult due to dependency induced by the internal feedback loops.  The conventional approach uses independent multiple training streams that employs plural copies of the network \cite{chen2014efficient}.  However, this inter-stream parallelism demands huge memory, which is a serious bottleneck for GPU based implementations. 
 
In this paper, we propose a parallelization approach as well as the generalized RNN structure. For this purpose, we first develop training algorithms for the generalized RNNs. The training equations of conventional LSTM can be perfectly represented with the generalized equations. Then, the parallelization approach exposes single-stream parallelization (intra-stream parallelism) that does not increase the size of mini-batches as the conventional multi-stream parallelization (inter-stream parallelism).  Experimental results show that further speed-up can be achieved by combining the two parallelism.

This paper is organized as follows. The generalized LSTM-like RNN structure is proposed and its training equations are derived in Section~\ref{sec:general}. In Section~\ref{sec:parallel}, the intra-stream parallelism of the generalized RNNs is explored and combined with the conventional inter-stream parallelism. In Section~\ref{sec:results}, experimental results of the proposed approach on a GPU are presented, followed by concluding remarks in Section~\ref{sec:conclusion}.

\section{Generalization}
\label{sec:general}
To apply our parallelization approach to various types of RNNs, we first introduce a generalized RNN structure that can represent complex RNNs using simple basic blocks. This generalization fully covers advanced LSTM network structures with forget gates and peephole connections, and their BPTT-based training algorithm. Also, with the generalized RNN, one can easily design a new RNN structure quite easily since every equation and the parallelization approach remain the same.

\subsection{Generalized RNN structure}

The proposed generalized RNN structure is basically a directed graph, which consists of a set of nodes and edges. Each node represents a layer and each edge makes a connection between two layers. There are two types of connections: delayed or not. A delayed connection makes a fixed amount of delay on the signal, and is used to construct a recurrent loop. More specifically, the connection $m$ propagates the activation of the source layer $k$ at the frame $t-d_m$ to the destination layer at the frame $t$ as
\begin{align}
\mathbf z_m(t) = \mathbf W_m \mathbf y_k(t - d_m),
\end{align}
where $\mathbf z_m$ is the output of the connection $m$, $W_m$ is the corresponding weight matrix, $\mathbf y_k$ is the activation of the source layer $k$, and $d_m$ is the amount of delay at the connection $m$. The value of $d_m$ is $0$ for non-delayed connections and larger than $0$ for delayed connections.

In an additive layer, the inputs are summed up and the activation function is applied on it:
\begin{align}
\mathbf s_k(t) &= \sum_{m \in A_k}{\mathbf z_m(t)}\\
\mathbf y_k(t) &= f_k(\mathbf s_k(t)),
\end{align}
where $\mathbf s_k$ is the state (input), $A_k$ is the set of the indices of the anterior connections, $\mathbf y_k$ is the activation, and $f_k(\cdot)$ is the activation function of the layer $k$. In addition to the normal additive layers, multiplicative layers are employed to represent gate units of LSTM networks. A multiplicative layer performs element-wise multiplication of input vectors (or matrices for batched computation) as follows:
\begin{align}
s_{k, i}(t) = \prod_{m \in A_k}{z_{m, i}(t)},
\end{align}
where the subscript $i$ represents the index of elements in a vector.

For generality, we introduce an aggregation function $g_k(\cdot)$ as
\begin{align}
\mathbf s_k(t) = g_k(\{ \mathbf z_m(t) | m \in A_k\}),
\end{align}
where $g_k(\cdot)$ is a vector addition function for an additive layer or an element-wise multiplication function for a multiplicative layer, or it can be other nonlinear functions to add further nonlinearity to the network.

In the previous approach on the generalized LSTMs \cite{monner2012generalized}, the gate units are implemented with gated connections. However, the gated connection has two input layers, so cannot be regarded as an edge of a familiar directed graph structure, where each edge has one input and one output.

\begin{figure}[!t]
	\centerline
	{%
		\includegraphics[width=2.9in]{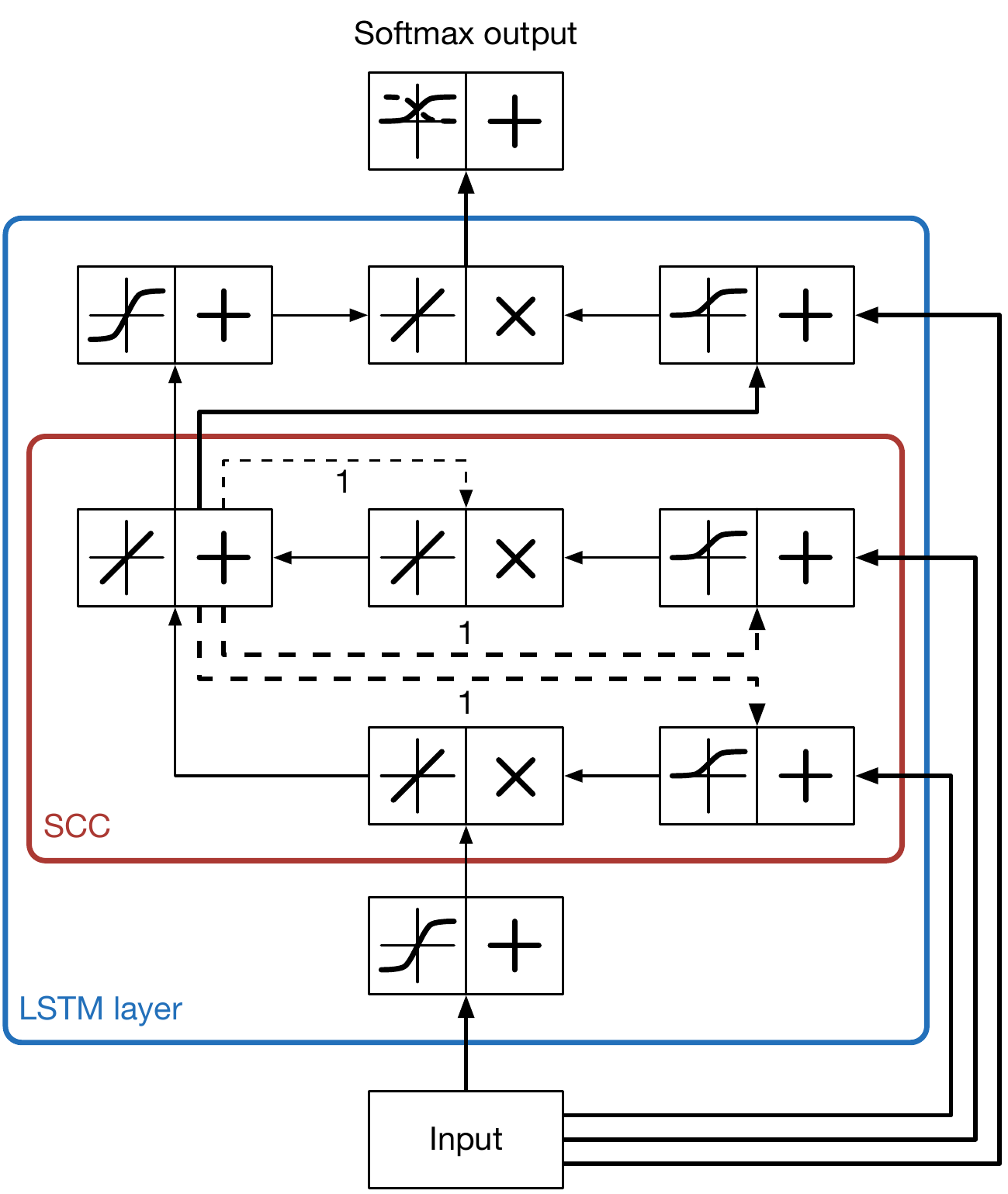}%
	}
	\caption{Generalized representation of an LSTM network with forget gates and peephole connections. Thick arrows represent connections with full weight matrices. On the other hand, connections with the thin arrows have identity weight matrices. The numbers on the dashed lines indicate the corresponding delay amounts. A non-singleton strongly connected component (SCC) is drawn, of which nodes will be grouped into a single recurrent node to make the network acyclic.}
	\label{fig:LSTM}
\end{figure}

In our approach, by introducing the multiplicative layers, LSTM gates can be regarded as normal nodes in a graph structure, which allows general graph algorithms to be directly applied in Section~\ref{sec:parallel}. As an example, \figurename~\ref{fig:LSTM} shows a generalized representation of a single-layer LSTM network with forget gates and peephole connections.

\subsection{Training}

In this section, BPTT \cite{werbos1990backpropagation} based training equations for the generalized RNN are derived. The objective is to minimize the following total error from $t_0 + 1$ to $t_1$:
\begin{align}
E^{\mathrm total}(t_0,t_1) = \sum_{t_0 < t \leq t_1}{E(t)},
\end{align}
where $E(t)$ is the error at frame $t$. For convenience, we define two derivative variables as
\begin{align}
\delta_{k, i}(t) &= -\frac{\partial E^{\mathrm total}(t_0,t_1)}{\partial s_{k, i}(t)} \\
\epsilon_{m, i}(t) &= -\frac{\partial E^{\mathrm total}(t_0,t_1)}{\partial z_{m, i}(t)}.
\end{align}
These two variables will be back-propagated at the backward pass. If the layer $k$ is an output layer, $\delta_{k, j}(t)$ should be initialized by comparing the output with a desired output $d_{k,j}(t)$ according to the error criterion defined by $E(t)$ and the activation function of the output layer. Using the minimum cross-entropy criterion with the softmax activation function,
\begin{align}
\delta_{k, j}(t) &= d_{k, j}(t) - y_{k, j}(t).
\end{align}
If the layer $k$ is not an output layer,
\begin{align}
\delta_{k, j}(t) &= -\sum_{n \in P_k}{ \sum_{i \in I_n}{ \frac{\partial E^{\mathrm total}(t_0,t_1)}{\partial z_{n, i}(t + d_n)} \frac{\partial z_{n, i}(t + d_n)}{\partial y_{k, j}(t)} }} \frac{\partial y_{k, j}(t)}{\partial s_{k, j}(t)} \\
&= \sum_{n \in P_k}{ \sum_{i \in I_n}{ \epsilon_{n, i}(t + d_n) W_{n, ij} }} f_k'( s_{k, j}(t) ) ,
\label{eq:delta}
\end{align}
where $P_k$ is the set of posterior connection indices of the layer $k$ and $I_n$ is the set of element indices of the vector $z_n$. Also, $\epsilon_{m, j}(t)$ becomes
\begin{align}
\epsilon_{m, j}(t) &= -\frac{\partial E^{\mathrm total}(t_0,t_1)}{\partial s_{k, j}(t)} \frac{\partial s_{k, j}(t)}{\partial z_{m, j}(t)} \\
&= \delta_{k, j}(t) \frac{\partial}{\partial z_{m, j}(t)} g_k(\{ \mathbf z_n(t) | n \in A_k\}),
\label{eq:epsilon}
\end{align}
where $k$ is the index of the destination layer of the connection $m$.
To truncate errors at $t = t_0'$, we backpropagate the two derivative variables while $t > t_0'$ where $t_0' \leq t_0$ using \eqref{eq:delta} and \eqref{eq:epsilon}. After the backward pass, the truncated error gradient of the connection $m \in P_k$ can be acquired by
\begin{align}
\frac{\partial E^{\mathrm total}(t_0,t_1)}{\partial W_{m, ij}} &\approx \sum_{t_0' < t \leq t_1}{\frac{\partial E^{\mathrm total}(t_0,t_1)}{\partial z_{m, i}(t)} \frac{\partial z_{m, i}(t)}{\partial W_{m, ij}}} \\
&= -\sum_{t_0' < t \leq t_1}{\epsilon_{m, i}(t) y_{k, j}(t - d_m)}.
\end{align}
In matrix form, \eqref{eq:delta} can be represented as
\begin{align}
\boldsymbol \delta_k(t) &= \bigg( \sum_{n \in P_k}{ \mathbf W_n^T \boldsymbol \epsilon_n(t + d_n) } \bigg)  \circ f_k'(\mathbf s_k (t)),
\end{align}
where $\circ$ denotes element-wise vector multiplication. If the layer $k$ is an additive layer, then \eqref{eq:epsilon} becomes
\begin{align}
\boldsymbol \epsilon_m(t) &= \boldsymbol \delta_k(t).
\end{align}
Otherwise for the multiplicative layer $k$,
\begin{align}
\boldsymbol \epsilon_m(t) &= \boldsymbol \delta_k(t) \circ \prod_{\mathclap{n \in A_k,n \neq m}}^{\circ}{\mathbf z_n(t)},
\end{align}
where element-wise multiplications are performed with $\prod$. The error gradient matrix for the connection $m \in P_k$ is computed by
\begin{align}
\nabla \mathbf W_m = - \sum_{t_0' < t \leq t_1} {\boldsymbol \epsilon_m(t) \mathbf y_k^T(t-d_m)}.
\end{align}
The error gradients can be used for the first order optimization methods such as stochastic gradient descent.

\section{Parallelization}
\label{sec:parallel}

Parallelization of RNN computation is quite challenging due to dependencies between two consecutive frames. The state of an RNN of the frame $k$ cannot be determined until the computation for the frame $k - 1$ is finished. In this section, we first develop a parallelization method for the forward and the backward pass with a single stream (intra-stream parallelism), and then extend the approach to a multi-stream case (inter-stream parallelism).

\subsection{Intra-stream parallelism}

The key concept of separating sequential parts from the parallel parts of an RNN is to determine loops in the RNN and group each loop into a single special node called a \emph{recurrent node}. Then, the remaining structure becomes a directed acyclic graph (DAG), which can be easily parallelized as in a mini-batch based feed-forward neural network computation. Only the internal computations of the recurrent nodes are performed sequentially.

\begin{figure}[!t]
	\centerline
	{%
		\includegraphics[width=1.7in]{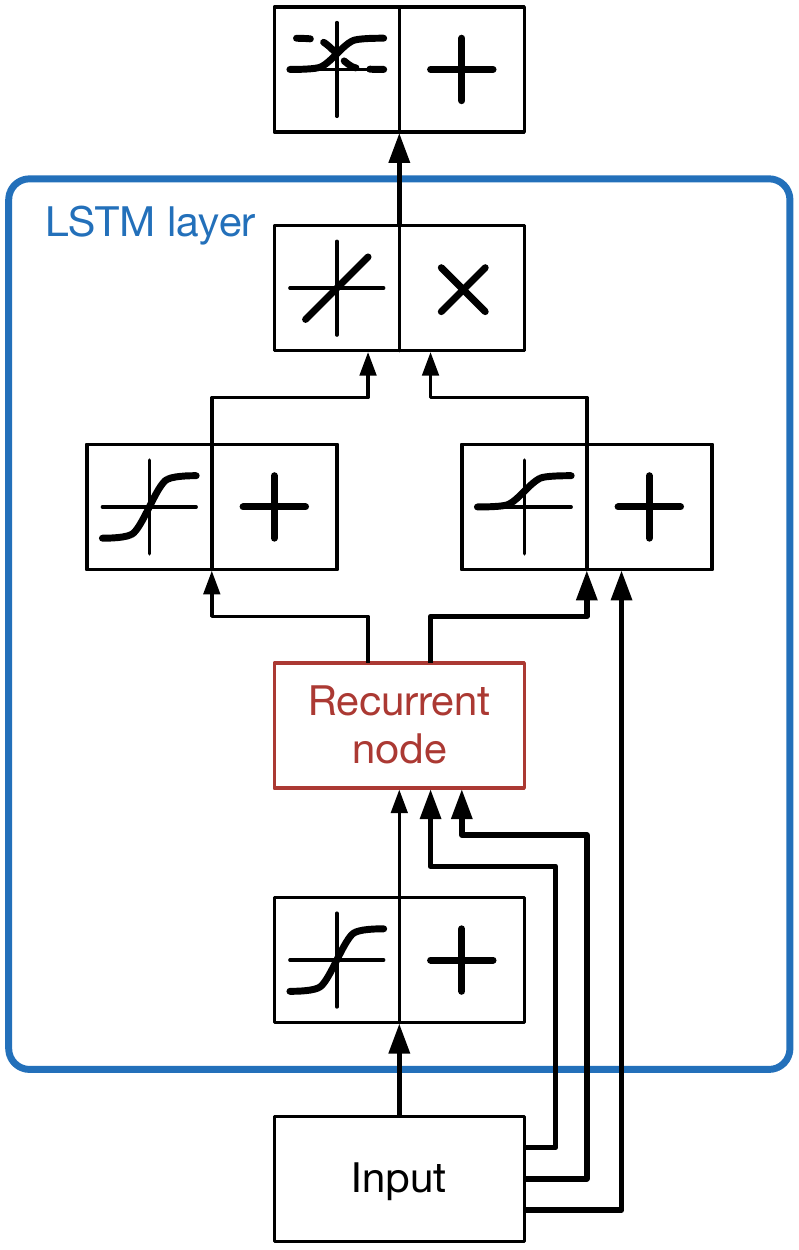}%
	}
	\caption{Feed-forward representation of the LSTM network that is depicted in \figurename~\ref{fig:LSTM}.}
	\label{fig:LSTM_2}
\end{figure}

More specifically, strongly connected components (SCCs) are found to determine which nodes should be grouped into a recurrent node. An SCC is a subgraph that is strongly connected, that is, there are one or more paths between every pair of two vertices inside the subgraph. An SCC analysis finds a set of SCCs that form a partition of the vertex set of the original graph. For SCCs that are singletons and do not contain a self-loop, the original nodes inside the SCCs remain unchanged. Otherwise, the nodes in each SCC are grouped into a single recurrent node. Then, the final graph becomes a DAG and be ready for parallel computation. An example of an LSTM network is shown in \figurename~\ref{fig:LSTM_2}. One of the famous algorithms for finding SCCs is the Tarjan's strongly connected component algorithm \cite{tarjan1972depth}. Tarjan's algorithm also provides a reverse topological sort of the resulting DAG, which is useful to determine the activation order.

Once an RNN is represented as a DAG, the forward computation becomes very similar to that of feedforward networks. As in the case of feedforward networks, computations of nodes and edges are performed in a topological order of the DAG. These operations can be done in parallel over several frames since the network is represented as a DAG and there are no dependencies between different frames except the isolated recurrent nodes.

Recurrent nodes are subgraphs of the original RNN and should be computed sequentially. The computation of a recurrent node from frame $t_0$ to $t_1$ in the forward pass requires $t_1 - t_0 + 1$ sequential steps. In each step of the forward pass, delayed connections are computed first. Then the remaining part excluding the delayed connections becomes a DAG and can be computed in a topological order. The computation of a backward pass can be performed similarly with reversed topological orders.

The sequential computations of recurrent nodes are quite expensive and often become a bottleneck of the overall performance. To speed up these sequential parts, we need to employ the multi-stream parallelization.

\subsection{Inter-stream parallelism}
\label{ssec:multi-stream}

Inter-stream parallelism can be explored in the multi-stream mode where an RNN processes $N$ streams with independent contexts. This is equivalent to running $N$ independent copies of the RNN. Therefore, the multi-stream mode greatly increases parallelism and the overall execution speed. Recently, this approach was successfully applied to speed up language model training with an Elman network on a GPU \cite{chen2014efficient}.

For training an RNN in the multi-stream mode, the input and target streams are usually given by connecting randomly ordered training sequences. Since the lengths of the training sequences are very long, we apply the efficient version of truncated BPTT($h$), denoted as BPTT($h; h'$) proposed in \cite{williams1990efficient}. BPTT($h; h'$) is similar to the ordinary truncated BPTT($h$) in that the network is unrolled $h$ times. However, in the forward pass of BPTT($h; h'$), $h'$ time steps are computed at once. Also, the error gradients for the recent $h'$ output errors are obtained by one iteration. These error gradients are summed up over the $N$ training streams. Therefore, output errors of total $N \times h'$ frames affect the error gradients when updating weights after backward passes. We call the set of these frames as a \emph{mini-batch} throughout the paper, as it is equivalent to a mini-batch in stochastic gradient descent methods of feedforward neural networks.

Increasing $N$ also speeds up the training. However, we cannot make $N$ very large since the size of a mini-batch, $N \times h'$, is limited by the physical memory size of a GPU. Moreover, increasing the size of a mini-batch results in infrequent update of the weights and may slow down the convergence \cite{byrd2012sample}. Also, the parameter $h'$ cannot be easily modified since the training speed is approximately proportional to the ratio of $h'$ to $h$. For simplicity, let us assume $h=2h'$ to fix the training speed. In this case, error propagates through $h'$ to $2h' - 1$ previous time steps in backward pass. Therefore $h'$ should be set sufficiently large to solve long time lag problems.

\section{Experimental Results}
\label{sec:results}

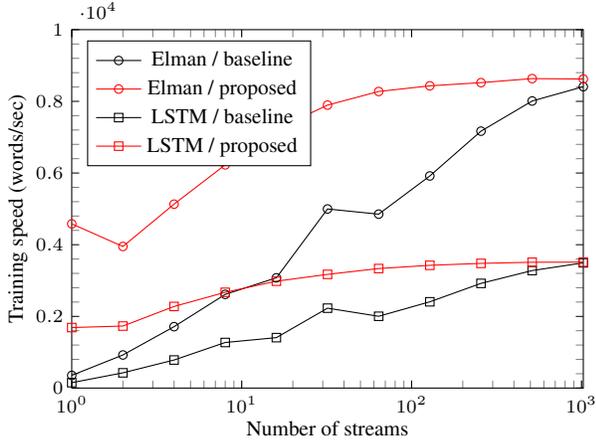
\begin{figure}[t!]
\centering
\centerline{%
\begin{tikzpicture}
\begin{axis}
[
width=3.3in,
height=2.5in,
compat=1.3,
xmin=1,
ymin=0,
xmax=1024,
ymax=10000,
label style={font=\footnotesize},
xlabel={Number of streams},
ylabel={Training speed (words/sec)},
xlabel shift=-3pt,
ylabel shift=-3pt,
legend style={font=\footnotesize,at={(0.95,0.95)}},
tick label style={font=\scriptsize},
domain=1:512,
legend style={at={(0.03, 0.97)}, anchor=north west},
minor x tick num=4,
minor y tick num=4,
xmode=log,
log basis x={10},
xtick pos=both,
xtick align=inside,
major tick style={line width=0.010cm, black},
major tick length=0.10cm
]%
\legend{Elman / baseline, Elman / proposed, LSTM / baseline, LSTM / proposed};
\addplot[color=black, mark=o, mark size=1.6pt, solid, mark repeat=1,mark options=solid]
file{data/LM_Elman_seq.txt};
\addplot[color=red, mark=o, mark size=1.6pt, solid, mark repeat=1,mark options=solid]
file{data/LM_Elman_par.txt};
\addplot[color=black, mark=square, mark size=1.6pt, solid, mark repeat=1,mark options=solid]
file{data/LM_LSTM_seq.txt};
\addplot[color=red, mark=square, mark size=1.6pt, solid, mark repeat=1,mark options=solid]
file{data/LM_LSTM_par.txt};
\end{axis}%
\end{tikzpicture}%
}%
\caption{Comparison of language model training speeds with Elman and LSTM networks. The LSTM employs forget gates and peephole connections. The sizes of the input layer, hidden or LSTM layer, and output layer is 38,000, 512, and 20,000 respectively. The mini-batch size is fixed to 1,024, so the error propagates from 1,024$/N$ to 2,048$/N - 1$ previous steps where $N$ is the number of streams. }
\label{fig:LM}
\end{figure}

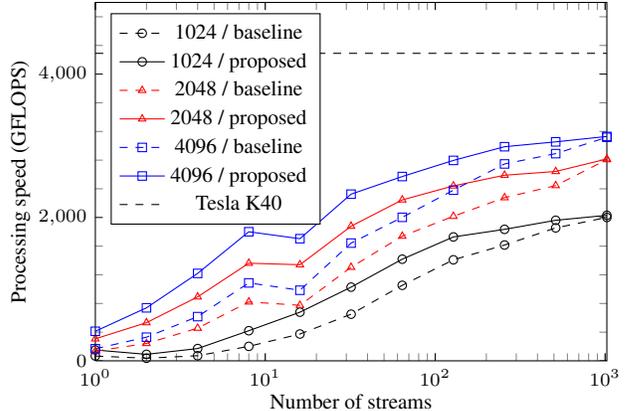
\begin{figure}[t!]
\centering
\centerline{%
\begin{tikzpicture}
\begin{axis}
[
width=3.3in,
height=2.5in,
compat=1.3,
xmin=1,
ymin=0,
xmax=1024,
ymax=5000,
label style={font=\footnotesize},
xlabel={Number of streams},
ylabel={Processing speed (GFLOPS)},
xlabel shift=-3pt,
ylabel shift=-3pt,
legend style={font=\footnotesize,at={(0.95,0.95)}},
tick label style={font=\scriptsize},
domain=1:256,
legend style={at={(0.03, 0.97)}, anchor=north west},
minor x tick num=4,
minor y tick num=4,
xmode=log,
log basis x={10},
xtick pos=both,
xtick align=inside,
major tick style={line width=0.010cm, black},
major tick length=0.10cm
]%
\legend{1024 / baseline, 1024 / proposed, 2048 / baseline, 2048 / proposed, 4096 / baseline, 4096 / proposed, Tesla K40};
\addplot[color=black, mark=o, mark size=1.6pt, dashed, mark repeat=1,mark options=solid]
file{data/LSTM_1024.txt};
\addplot[color=black, mark=o, mark size=1.6pt, solid, mark repeat=1,mark options=solid]
file{data/LSTM_1024_par.txt};
\addplot[color=red, mark=triangle, mark size=1.6pt, dashed, mark repeat=1,mark options=solid]
file{data/LSTM_2048.txt};
\addplot[color=red, mark=triangle, mark size=1.6pt, solid, mark repeat=1,mark options=solid]
file{data/LSTM_2048_par.txt};
\addplot[color=blue, mark=square, mark size=1.6pt, dashed, mark repeat=1,mark options=solid]
file{data/LSTM_4096.txt};
\addplot[color=blue, mark=square, mark size=1.6pt, solid, mark repeat=1,mark options=solid]
file{data/LSTM_4096_par.txt};
\addplot[color=black, mark=none, mark size=1.6pt, dashed, mark repeat=1,mark options=solid]
file{data/K40.txt};
\end{axis}%
\end{tikzpicture}%
}%
\caption{Comparison of GPU processing power utilizations when training LSTM networks with the three different sizes of LSTM layers: 1,024, 2,048, and 4,096. The input and output layers have the same size as the LSTM layer. Also, the theoretical peak performance of Tesla K40 GPU is shown. The mini-batch size is fixed to 1,024.}
\label{fig:LSTM_size}
\end{figure}

Nvidia Tesla K40 GPU is used for the following experiments. For all experiments, BPTT($2h$; $h$) is used for simplicity. Since the training algorithm for the generalized RNN structure is mathematically equivalent to that of Elman or LSTM networks, results with performance measures such as accuracy or the mean squared error (MSE) are not reported.

To evaluate the proposed parallelization approach, we evaluate the language model training speed with the multi-stream mode as in \cite{chen2014efficient}. The RNN architecture is an Elman network with 38,000 input, 512 hidden, and 20,000 output units. The mini-batch size is fixed to 1,024 to use the same amount of GPU memory. Hence, with $N$ streams, $h=1$,024$/N$ and the error propagates from 1,024$/N$ to 2,048$/N - 1$ previous time steps. For comparison, an LSTM version of the network with forget gates and peephole connections are also evaluated. Note that the LSTM network has no self recurrent connection from the output of the LSTM layer to the input of that.

The training speeds are compared in \figurename~\ref{fig:LM} with varying number of streams. Since the baseline approaches does not exploit intra-stream parallelism, they show poor training speeds when the number of streams are small. On the other hand, the proposed approach employs intra-stream parallelism and shows over 10 times of speed-up over the baseline approach when a single stream is used. Also, with the proposed approach, we can obtain almost the maximum speed only with 64 streams. This is a nice advantage since using less number of streams allows RNNs to learn longer time lags when the size of mini-batch is limited, as discussed in Section~\ref{ssec:multi-stream}.

To analyze scalability and GPU efficiency with various size of networks, we perform another experiment with LSTM networks with forget gates and peephole connections. All layers of each network have the same size, which is 1,024, 2,048, or 4,096. To examine the GPU utilizations, we present the number of single-precision floating point operations per second (FLOPS) in \figurename~\ref{fig:LSTM_size} along with the theoretical peak performance of Tesla K40 GPU. Note that only the operations for parameters and error gradients are counted. Compared to the previous experiment where the input and output layers are very large, this example is much closer to the deep RNN architectures in terms of the ratio of the sequential computations (inside the recurrent nodes) to the parallel computations. As shown in the figure, the GPU utilization gets higher as the layer size or the number of streams increases. Also, the intra-stream parallelism further accelerates the training especially with the small number of streams.

\section{Concluding Remarks}
\label{sec:conclusion}

We introduced a generalized structure for RNNs which covers LSTM networks with forget gates and peephole connections. This generalized structure is represented as a directed graph where nodes and edges correspond to layers and connections, respectively. Due to the graph representation, we can automatically find loops inside RNNs using the Tarjan's strongly connected component algorithm and explore intra-stream parallelism. The proposed intra-stream parallelism is combined with inter-stream parallelism in multi-stream mode for further acceleration. The experiments show that exploiting these two parallelisms greatly speeds up the training task on a GPU.

\pagebreak

\bibliographystyle{IEEEbib}
\bibliography{refs}

\end{document}